\documentclass[10pt,twocolumn,letterpaper]{article}

\usepackage[pagenumbers]{iccv} 
\usepackage{multirow}

\usepackage{caption}
\usepackage{cuted}  
\usepackage{fancyhdr}

\definecolor{iccvblue}{rgb}{0.21,0.49,0.74}
\usepackage[pagebackref=false,breaklinks=true,colorlinks,citecolor=blue,urlcolor=blue,linkcolor=blue,bookmarks=false]{hyperref}
\usepackage{bbding}

\def\confName{ICCV}
\def\confYear{2025}

\title{Pixel-SAIL: Single Transformer For Pixel-Grounded Understanding}

\author{
\centerline{
Tao Zhang\textsuperscript{\rm 1,2} \qquad
Xiangtai Li\textsuperscript{\rm 1}  \qquad
Zilong Huang\textsuperscript{\rm 1}  \qquad
Yanwei Li\textsuperscript{\rm 1}  \qquad
Weixian Lei\textsuperscript{\rm 1}  \qquad
} \\
\centerline{
Xueqing Deng\textsuperscript{\rm 1} \qquad 
Shihao Chen\textsuperscript{\rm 2} \qquad
Shunping Ji\textsuperscript{\rm 2} \qquad
Jiashi Feng\textsuperscript{\rm 1} \qquad
}\\
\centerline{
\textsuperscript{\rm 1} Bytedance Seed \quad
\textsuperscript{\rm 2} WHU
} \\
\centerline{
\normalsize Project Page: \url{https://zhang-tao-whu.github.io/project/pixelsail} 
}
}

\begin{document}

\maketitle

    

\captionsetup{hypcap=false}
\begin{strip}
    \vspace*{-40pt} 
    \centering
    \includegraphics[width=\linewidth]{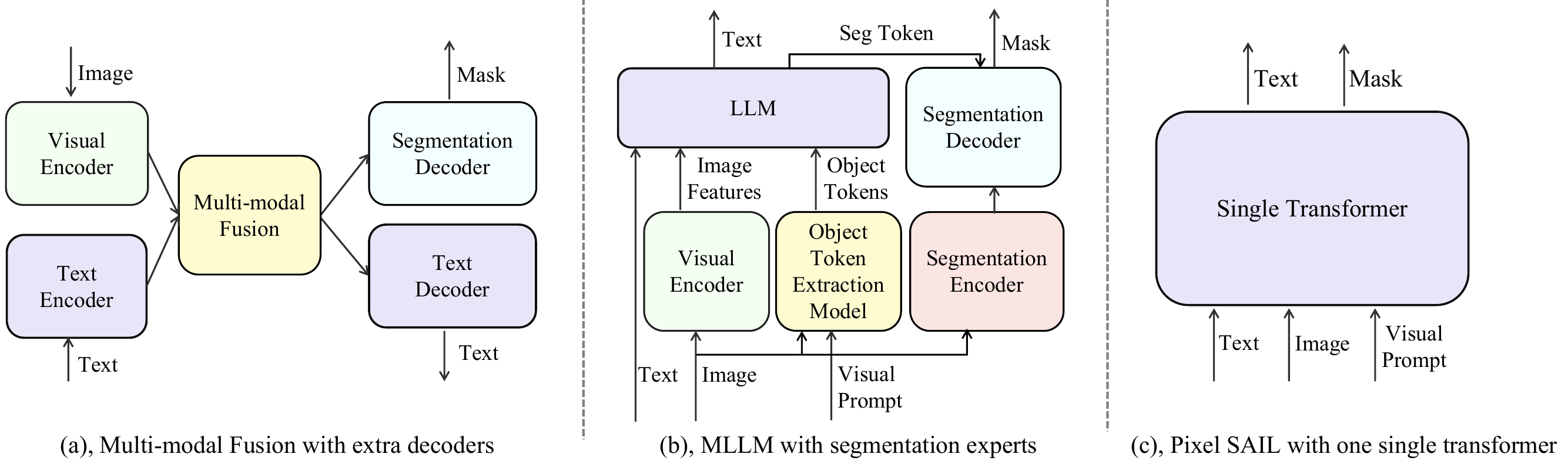}
    \captionof{figure}[teaser]{\small \textbf{Comparison of current MLLMs for pixel-wise understanding with our method.} (a) and (b). Current MLLMs for pixel-wise understanding feature highly complex system architectures, including an LLM, a CLIP-like vision backbone, an object token extraction model, a segmentation vision backbone, and a SAM-like decoder. (c). Our method employs only a single transformer.}
    \vspace{-5pt}
    \label{fig:teaser}
\end{strip}

\thispagestyle{fancy}
\fancyhead[L]{\includegraphics[height=15pt]{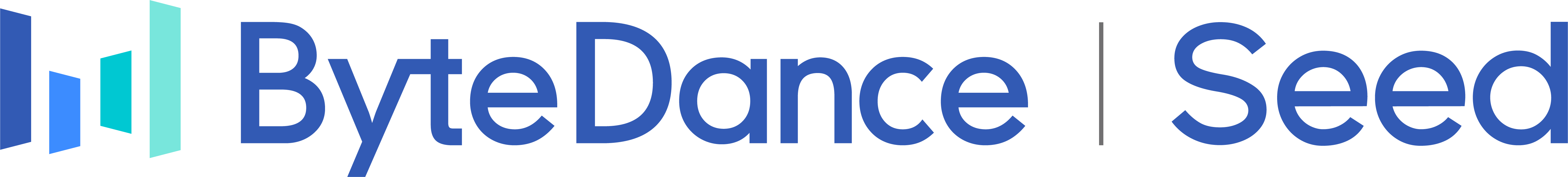}}
\fancyhead[R]{}

\begin{abstract}
Multimodal Large Language Models (MLLMs) achieve remarkable performance for fine-grained pixel-level understanding tasks. 
However, all the works rely heavily on extra components, such as vision encoder (CLIP), segmentation experts, leading to high system complexity and limiting model scaling. 
In this work, our goal is to explore a highly simplified MLLM without introducing extra components. 
Our work is motivated by the recent works on Single trAnsformer as a unified vIsion-Language Model (SAIL) design, where these works jointly learn vision tokens and text tokens in transformers. 
We present Pixel-SAIL, a single transformer for pixel-wise MLLM tasks. In particular, we present three technical improvements on the plain baseline. First, we design a learnable upsampling module to refine visual token features. 
Secondly, we propose a novel visual prompt injection strategy to enable the single transformer to understand visual prompt inputs and benefit from the early fusion of visual prompt embeddings and vision tokens.
Thirdly, we introduce a vision expert distillation strategy to efficiently enhance the single transformer's fine-grained feature extraction capability.
In addition, we have collected a comprehensive pixel understanding benchmark (PerBench), using a manual check.
It includes three tasks: detailed object description, visual prompt-based question answering, and visual-text referring segmentation. 
Extensive experiments on four referring segmentation benchmarks, one visual prompt benchmark, and our PerBench show that our Pixel-SAIL achieves comparable or even better results with a much simpler pipeline.
Code and model will be released at \url{https://github.com/magic-research/Sa2VA}.
\end{abstract}    
\vspace{-6mm}\section{Introduction}
\label{sec:intro}








Multi-modal Large Language Models (MLLMs) have garnered significant research efforts, driven by advancements of Large Language Models (LLMs)~\cite{llama,llama2, qwen2}. 
While most studies focus on open-ended visual question answering tasks, there is a growing interest~\cite{hanoona2023GLaMM,OMGLLaVA} in fine-grained, pixel-level understanding. 
This enables broader applications, such as facilitating precise region-level editing and generation and achieving precise understanding of designated mask regions.

Recent pixel-wise MLLMs~\cite{lai2023lisa, ren2023pixellm, xia2023gsva, zhang2024groundhog, hanoona2023GLaMM, OMGLLaVA, yuan2025sa2va} mainly adopt visual and language fusion frameworks, following design patterns~\cite{liu2023polyformer, ding2021vision, yang2022lavt} established before the LLM era. 
For example, LAVIT~\cite{yang2022lavt} adopts encoder-fusion approach, injecting language embedding (generated by BERT~\cite{devlin2019bert}) into vision transformers.
With the advent of LLMs~\cite{llama2, qwen2, yang2024qwen2}, recent works~\cite{lai2023lisa, ren2023pixellm, OMGLLaVA, yuan2025sa2va} integrate state-of-the-art segmentation models~\cite{kirillov2023segment, ravi2024sam2, OMGSeg}, for pixel-level understanding, by either appending them to LLM outputs or embedding LLM within segmentation pipelines.
While effective, the overall architectures are complex, requiring specialized components such as vision-language fusion modules and additional decoders. 
Moreover, their final performance often heavily depends on either MLLMs or the segmentation models, which may lead to suboptimal results due to limitations within individual submodules.

In this work, we explore a novel, simple yet effective pixel-wise MLLM design, drawing inspiration from recent advancements in SAIL architecture, which is also called Encoder-free MLLMs.
These methods drop the extra vision encoder and jointly co-train vision and language tokens on large scale datasets, with a simpler design. 
Moreover, they show competitive performance on image-level VQA tasks, compared with LLaVA.
Motivated by this success, we extend the framework to pixel-level understanding tasks, aiming to reduce the complexity of existing approaches.
%
To the best of our knowledge, this is the first study to explore the simplest architecture for pixel-wise MLLM tasks, including referring segmentation and visual prompt understanding.

We first directly extend SAIL architecture by adding segmentation token and visual prompt tokens to generate segmentation masks and output region caption, following previous works~\cite{hanoona2023GLaMM,yuan2023osprey, lai2023lisa}.
However, this leads to inferior results on both segmentation and visual prompt understanding.
Several reasons are: (1), The misalignments on high resolution features since there are no segmentation decoders since SAIL directly reshape the vision tokens into features.
(2), Previous works directly adopt mask pooling on high level visual tokens where SAIL baseline only maps RGB inputs with one projection layer, where most tokens are low level features.
(3), The mask quality is low since no segmentation experts are involved.

To solve these problems, we present three simple technical improvements, which lead to our Pixel-SAIL framework. First, we design a simple learnable up-sampling module to refine the low resolution visual tokens in high resolution features. 
Our goal is to keep the design as simple as possible, where only one transposed 2D convolution is involved.
Then, for visual prompt understanding, we design a novel visual prompt injection method, where we map the visual prompts into special text tokens without introducing extra visual prompt encoder in the middle stage of SAIL.
Next, we propose to distill the previous segmentation experts into SAIL to improve mask quality.
All the improvements are plug-in-play, and we verify the effectiveness on various SAIL architectures, including SOLO~\cite{chen2024single} and EVEv2~\cite{diao2025evev2}.

Then, to further indicate the effectiveness of our Pixel-SAIL and facilitate the development of pixel-LLM community, we further design a new challenging benchmark, PerBench. Compared with previous pixel-wise MLLM benchmarks, we have three innovative and challenging features. 
First, we include a detailed object caption where most existing benchmarks only contain short captions without fine-gained contents. 
Secondly, we re-evaluate visual-prompt understanding as multi-choice VQA tasks following MME~\cite{fu2023mme} and MMBench~\cite{liu2023mmbench} to achieve more accurate region caption evaluation.
Thirdly, we introduce a task by segmenting objects jointly referenced by visual prompts and text.
Our benchmark reveals the limitation of current state-of-the-art pixel-wise MLLM on fine-grained understanding and mixed referring tasks.

Pixel-SAIL is jointly co-trained with mixed data engine on referring segmentation datasets, VQA datasets, and visual prompt datasets. 
Experimental results show that our method can achieve better results on five pixel-wise benchmarks. 
In particular, on RefCOCOg and RefCOCO+ datasets, our method with 3B size can outperform previous pixel MLLMs, including GLaMM (7B) and OMG-LLaVA (7B), by 1.5-3.0\%, with a simpler pipeline.
On our PerBench, our method achieves 24.2 METEOR, 74\% accuracy, 33.4 cIoU and 42.2 overall score, surpassing the SOTA MLLMs GLaMM (7B) and Sa2VA (4B) with overall scores of 26.9 and 3.2, respectively.


\section{Related Work}
\label{sec:related_work}
%
%
%

\noindent
\textbf{Large Vision Language Models.} Staring from CLIP~\cite{radford2021learning} and ALIGN~\cite{jia2021scaling}, modern vision language models have adopted contrastive learning on large-scale image-text datasets for learning vision-text aligned representations.
The trained models are also proven to work well on open-vocabulary perception, such as segmentation~\cite{yuan2024ovsam,luddecke2022image, zhang2023dvis, zhang2023dvis++} and detection~\cite{gu2021open,wu2023open,wang2023v3det,zareian2021open}. 
The following works~\cite{li2022blip,zhai2023sigmoid,li2023blip,xu2023metaclip} share the same network design, exploring modified loss functions and targeting data quality and filtering. 
Then, with the rise of large language models~\cite{llama2,qwen2,llama,cai2024internlm2}, recent works~\cite{Qwen-VL,internlmxcomposer,chen2024internvl,liu2023llava,chen2024expanding,tong2024cambrian1} mainly focus on multi-modal large language models for open-ended settings, such as visual question answering or OCR benchmarks. 
On representative work, LLaVA~\cite{liu2023llava}, uses the CLIP to encode images into visual tokens and sends the visual tokens to LLMs. 
After that, the following works~\cite{Qwen-VL,liu2024llavanext,li2024llava_onevision} improve designs with scaled high quality datasets, images, and videos co-training. 
Meanwhile, several recent works~\cite{diao2025evev2,chen2024single,diao2024EVE,luo2024mono} also explore the visual encoder-free designs, which jointly learn the image and text representation in a single transformer architecture. 
For example, SOLO~\cite{chen2024single} collects mixed language and vision datasets and trains one transformer for VQA tasks, while EVE~\cite{diao2024EVE} designs a CLIP supervision to enhance visual token learning. 
Our work follows the visual encoder-free design, and we go a step further by exploring pixel-grounded understanding tasks, including grounding tasks and visual prompt understanding.
To our knowledge, we are the first to apply encoder-free architecture for pixel-grounded understanding tasks.

\noindent
\textbf{Referring Expression Segmentation.} This task outputs specific masks driven by text description. 
Earlier works~\cite{EFN,yang2021lavt,hui2020linguistic,liang2022local,GRES} explore various fusion architecture and modules to enhance text and vision feature alignments. 
Equipped with LLMs, several recent advanced works~\cite{lai2023lisa, OMGLLaVA, xia2023gsva,qi2024generalizable, yuan2025sa2va, yuan20254th, munasinghe2024videoglamm,hanoona2023GLaMM, zhou2025they} propose more complex referring tasks, including reasoning referring or joint mask and caption generation. 
In particular, LISA~\cite{lai2023lisa} involves complex expression while GLaMM~\cite{hanoona2023GLaMM} annotates a new dataset and proposes region-level caption and segmentation tasks. 
However, all these works contain complex designs: extra vision encoders, segmentation encoders, mask decoders, and prompt encoders. 
Our method, Pixel-SAIL, only has one transformer to jointly learn the joint visual and language feature. 
With proposed data engine and improved methods, Pixel-SAIL achieves good results with much simpler architecture.

\noindent
\textbf{Visual Prompt Understanding.} Understanding visual prompts plays an important role when building interaction between VLMs and human. 
Recent works~\cite{cai2024vipllava,yuan2023osprey,hanoona2023GLaMM,ma2024groma,lin2024draw} build new visual prompt datasets for region caption generation and prompt-aware VQA tasks. 
ViP-LLaVA~\cite{cai2024vipllava} overlays the visual prompts directly onto the image canvas and fine-tunes the LLaVA on a specific visual prompt dataset, while Osprey~\cite{yuan2023osprey} explores pixel-wise mask regions into language instructions.
Our method can also be extended into visual prompt understanding with our proposed prompt token injection design.

\section{Method}
\label{sec:method}
%




\begin{figure*}
  \centering
  \includegraphics[width=0.92\linewidth]{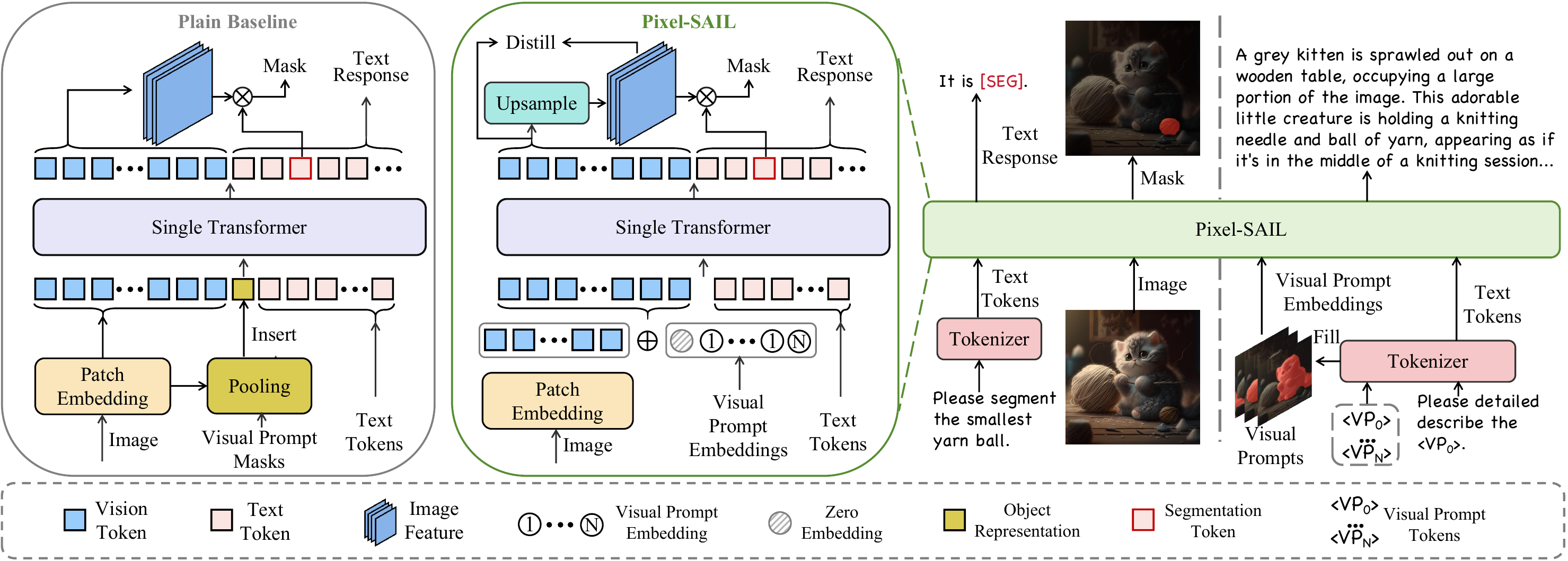}
  \caption{\textbf{The architecture of our proposed plain baseline and Pixel-SAIL.} Pixel-SAIL is as simple and elegant as the plain baseline but demonstrates significantly improved performance. The examples on the right demonstrate that Pixel-SAIL possesses the capability for general conversation and comprehensive pixel-grounded understanding.}\vspace{-2mm}
  \label{fig:archetecture}
\end{figure*}

\subsection{Encoder Free MLLM and Plain Baseline}
\label{sec:method_pre_knowledge}

Recently, several encoder-free MLLMs~\cite{chen2024single, luo2024mono, diao2025evev2, diao2025unveiling} achieve comparable performance with those extra vision encoders. These models jointly learn vision and text features in a single transformer, with much simpler architecture. In particular, SOLO uses a simple project layer to map the image into visual tokens and then combines language tokens as the inputs of the transformer.
However, no works have explored such new architecture for fine-grained vision language tasks (region caption, referring masks).

\noindent
\textbf{Plain Baseline.}
To fill this gap, we first construct a plain single transformer baseline, motivated by the previous ViT-based MLLMs~\cite{lai2023lisa,yuan2025sa2va}. 
We start it with a pre-trained encoder-free MLLM.
For segmentation tasks, we modify previous mask generation methods into the single transformer.
First, we reshape the hidden states of the last transformer layer of vision tokens $\mathcal{V}\in\mathbb{R}^{N \times C}$ into image features $\mathcal{F}\in\mathbb{R}^{\frac{H}{S} \times \frac{W}{S} \times C}$.
$N$ represents the number of vision tokens, $C$ denotes the channel size, $H$ and $W$ indicate the height and width of the image, $S$ stands for the downsampling stride.
Then, the image features are then cross-multiplied with the hidden states of the predicted segmentation token $\mathcal{Q}\in \mathbb{R}^{K \times C}$ to generate the segmentation masks $\mathcal{M}\in \mathbb{R}^{K \times \frac{H}{S} \times \frac{W}{S}}$. 
$K$ signifies the number of predicted segmentation tokens, following previous works~\cite{lai2023lisa,hanoona2023GLaMM}.
For visual prompt understanding, we employ a pooling-based method~\cite{yuan2023osprey} to derive object representations $\mathcal{O}\in \mathbb{R}^{M \times C}$ from image patch embeddings $\mathcal{P}\in \mathbb{R}^{\frac{H}{P} \times \frac{W}{P} \times C}$. 
These object embeddings are fed into the single transformer to represent the corresponding objects. $M$ represents the number of visual prompts, and $P$ denotes the patch size. 
For segmentation tasks, we adopt extra mask loss. Otherwise, we adopt the same text loss for VQA tasks and visual prompt understanding tasks.

\noindent
\textbf{Limitation.} The plain baseline demonstrates a certain level of pixel-text alignment capability since both segmentation token and visual prompt token are jointly learned with vision and language tokens.
However, the plain baseline exhibits several significant shortcomings:
1) The segmentation mask quality is poor due to the large feature down-sampling stride (16 or 32), even when using simple pixel shuffle or bilinear interpolation for up-sampling. 
2) The single transformer struggles to comprehend the referential target of object representation, as the object representation is summarized from image patch embeddings with poor semantic information.

\subsection{Pixel-SAIL Method}
\label{sec:stpu_method}

Given the substantial shortcomings, the performance of plain baseline in fine-grained pixel understanding tasks falls significantly, compared to vision-expert competitors (Sec.\ref{sec:exp}). 
To solve these challenges, we have implemented three key enhancements to the baseline architecture. First, we integrate a learnable up-sampling module to fully exploit the segmentation capabilities of the single transformer architecture. 
Second, we develop an innovative visual prompt injection mechanism that facilitates effective interpretation of visual prompt inputs.
Our method enables early-stage fusion between vision tokens and visual prompt embeddings. 
Finally, we introduce a dense feature distillation strategy that significantly improves the model's capacity for extracting fine-grained visual features. 
These improvements collectively address the shortcomings of the plain baseline while maintaining its architectural simplicity.

\noindent
\textbf{Learnable Up-sampling Module.} Inspired by~\cite{li2022exploring}, we also incorporate a simple learnable up-sampling model $\mathcal{U}$ to generate the high-resolution features $F_{h} \in \mathbb{R}^{\frac{H}{4}\times\frac{W}{4}\times C}$ essential for pixel-level grounding. 
The up-sampling module comprises multiple up-sampling blocks, each consisting of a transposed 2D convolution followed by a depth-wise convolution.
It effectively upscales the low-resolution features $F_{l} \in \mathbb{R}^{\frac{H}{S}\times\frac{W}{S}\times C} $, derived from resized vision tokens, to one-quarter of the original resolution.

\noindent\textbf{Visual Prompt Injection.} Previous works~\cite{yuan2023osprey, hanoona2023GLaMM, yuan2025sa2va} summarize the referenced object features via pooling on vision tokens from ViT encoder.
However, there are no such visual tokens for encoder-free MLLMs.
%
Thus, the inherent semantic deficiency hinders the single transformer's ability to precisely identify referenced objects based solely on feature summaries derived from patch embeddings, where most are low-level cues, such as edges.

To overcome this limitation, we propose an innovative visual prompt injection mechanism. 
Our approach integrates multiple visual prompt special tokens $\{VP_i | i \in [1, N]\}$ into the large language model's vocabulary. 
These tokens' text embeddings $\mathcal{VP}^{t}\in \mathbb{R}^{N\times C}$ are used to fill mask-based visual prompts $\mathcal{M}^{vp}\in \mathbb{R}^{N\times \frac{H}{P}\times \frac{W}{P}}$, thereby creating visual prompt tokens $\mathcal{VP}\in \mathbb{R}^{\frac{HW}{P^{2}}\times C}$. 
The vision tokens $\mathcal{V}\in \mathbb{R}^{\frac{HW}{P^{2}}\times C}$ are first added with these visual prompt tokens $\mathcal{VP}$ before being processed by the single transformer. 
This enhancement enables the model to accurately identify referenced objects by leveraging the corresponding special tokens $\{VP_i | i \in [1, N]\}$ within the text instructions.

\noindent\textbf{Dense Feature Distillation.} Due to the lack of large-scale, high-quality segmentation data like SA-1B~\cite{kirillov2023segment}, the method produces poor-quality masks, particularly at object boundaries. 
However, directly training on large-scale segmentation datasets would be costly and damage the original instruction following capabilities. 
To address both, we employ pre-trained segmentation experts to distill the single transformer, ensuring optimization of object details without hurting VQA capabilities.
We perform distillation by leveraging mask features generated by Mask2Former's~\cite{cheng2021mask2former} pixel decoder on the upsampled mask features $F_{h} \in \mathbb{R}^{\frac{H}{4}\times\frac{W}{4}\times C}$, and utilizing features produced by SAM2's~\cite{ravi2024sam2} encoder on the
low-resolution features $F_{l} \in \mathbb{R}^{\frac{H}{S}\times\frac{W}{S}\times C} $. 
This simple distillation strategy improves segmentation quality with only a negligible increase in training time.

\subsection{Benchmark and Dataset Engine}
\label{sec:stpu_training}

\begin{figure}
  \centering
  \includegraphics[width=1.0\linewidth]{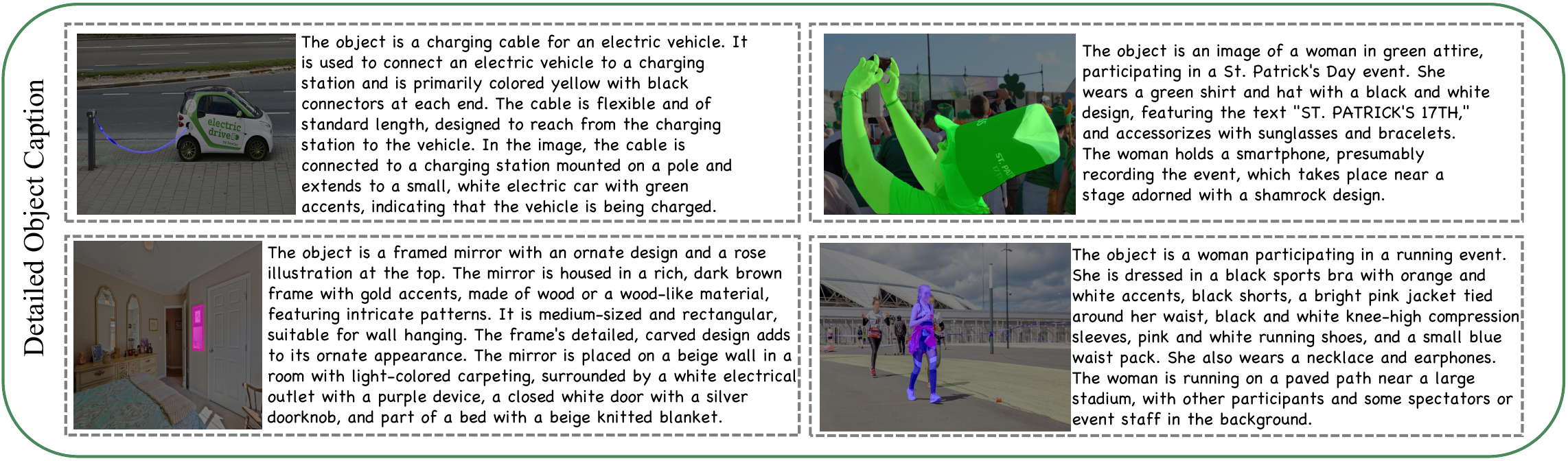}
  \includegraphics[width=1.0\linewidth]{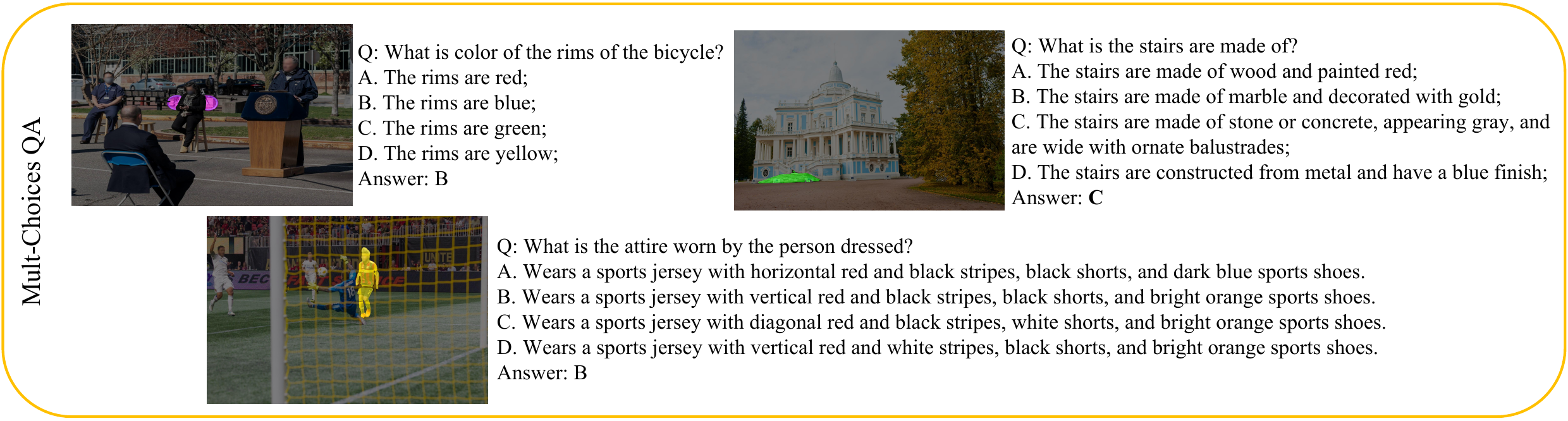}
  \includegraphics[width=1.0\linewidth]{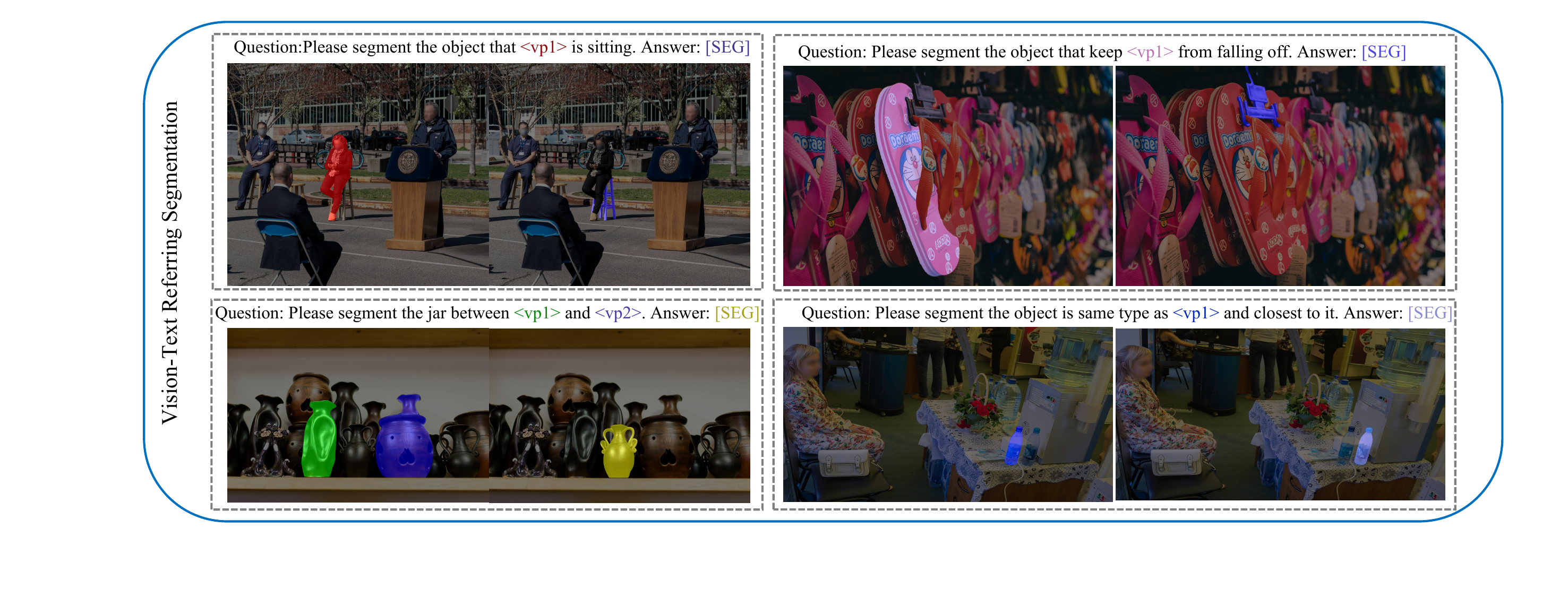}
  \caption{\textbf{Visual examples on our PerBench. Best view it in color and zoom in.}}\vspace{-5mm}
  \label{fig:benhcmark}
\end{figure}

\noindent
\textbf{Our Benchmark: PerBench.} 
%
%
We further manually annotate a benchmark named \textbf{PerBench} (\textbf{P}ix\textbf{e}l-grounded Unde\textbf{r}standing \textbf{Bench}mark). PerBench aims to address three aspects lacking in existing pixel grounding benchmarks.

The first aspect is detailed object caption. 
Previous works~\cite{chen2023sharegpt4v, li2024densefusion} have emphasized more detailed image captions, demonstrating that comprehensive captions significantly enhance model performance. 
However, current object caption datasets such as Osprey-724k~\cite{yuan2023osprey} and evaluation benchmarks like Refcocog provide only cursory object captions. 
To address this limitation, we leverage SOTA models InternVL2.5-78B~\cite{chen2024internvl} and Qwen2.5VL-72B~\cite{bai2025qwen2} to generate detailed object captions. These detailed object captions are then meticulously screened and refined through \textbf{manual review}, ultimately yielding 500 precise, nuanced object captions to serve as a robust evaluation benchmark. METEOR~\cite{banerjee2005meteor} serves as the evaluation metric for the detailed object caption task.

The second aspect is the assessment of visual-prompt understanding ability in multiple-choice format. 
Although captioning tasks can accurately reflect a model's visual prompt understanding ability, precise and fair evaluation is difficult. 
Rule-based metrics such as CIDEr~\cite{vedantam2015cider} and METEOR~\cite{banerjee2005meteor} are affected by response length, format, and ground-truth quality, while using models as evaluators inevitably introduces model bias. 
Therefore, a fair and quantitative visual-prompt understanding benchmark is necessary. 
Inspired by MMBench~\cite{liu2023mmbench} and MME~\cite{fu2023mme}, we manually annotated 500 multiple-choice questions based on detailed object captions, covering the examination of models' understanding of referenced objects' appearance, attributes, uses, and relationships with surrounding objects. 
MLLMs need to perceive the attributes of referenced objects accurately and have instruction-following ability to select the appropriate choice correctly. Accuracy is selected as the evaluation metric for the visual prompt-based multiple-choice questions.

The third aspect is segmenting objects jointly referenced by visual prompts and text, abbreviated as V-T RES. 
It aims to test the model's ability to understand objects indicated by user-input visual prompts and segment associated objects according to text instructions. 
This task comprehensively assesses the MLLM's pixel-grounded understanding ability, requiring the model to possess precise visual prompt understanding capabilities, text reasoning abilities, and pixel grounding skills. 
We also manually annotate 500 V-T RES samples, which five expert annotators double-check. Similar with RefCOCO series datasets, we select cIoU and gIoU as the evaluation metric for V-T RES task. The overall score of PerBench is the average of the normalized scores (0-100) from the above three tasks.

Our benchmark can be used to evaluate pixel-wise MLLMs and point out more challenging directions for detailed object understanding, joint visual prompts, and text understanding to the current community.

\noindent
\textbf{Dataset Engine.}
To fully unleash the potential of the single transformer, we collect diverse pixel-grounded data, including segmentation datasets and visual-prompt understanding datasets, following previous works~\cite{diao2025evev2,luo2024mono}.

For segmentation-related data, we first use RefCOCO/+/g~\cite{kazemzadeh2014referitgame, yu2016modeling} and COCO~\cite{coco_dataset} semantic segmentation data used in LISA~\cite{lai2023lisa}, the Grandf dataset (214k samples) used in GLaMM~\cite{hanoona2023GLaMM}, and MUSE data (246k samples) used in PixelLM~\cite{ren2023pixellm}. 
We also use recent Pixel2Cap~\cite{you2025pix2cap} data (comprising 20k images) and organized it into the referring segmentation format. 
Finally, we further add COCO~\cite{coco_dataset} panoptic segmentation data and structured it as: \textit{``Question: Please segment the \{class name\} in instance mode. Answer: \{class name\}-1 [SEG], ..., \{class name\}-n [SEG]."}

For visual prompt understanding, we employ two public datasets: Osprey-724k~\cite{yuan2023osprey} and Pixel2Cap~\cite{you2025pix2cap}. Additionally, we reformat the COCO dataset into a question-answer structure specifically designed to query object categories. 
To enhance the model's capability for fine-grained object description, we prompt the InternVL2.5-78B~\cite{chen2024internvl} model to generate approximately 300k detailed object captions derived from 10k SA-1B~\cite{kirillov2023segment} images.
Lastly, to maintain the instruction following ability, we also integrate the LLaVA-1.5~\cite{liu2023llava} 665k dataset into our training data.

\noindent
\textbf{Training.}
We combine all the aforementioned data for co-training. The loss function consists of the next token prediction loss $\mathcal{L}_{ntp}$, the segmentation loss $\mathcal{L}_{seg}$, and the distillation loss $\mathcal{L}_{distill}$:
\begin{equation}
    \mathcal{L} = \mathcal{L}_{ntp} + \mathcal{L}_{seg} + \alpha\mathcal{L}_{distill}, \quad \mathcal{L}_{seg} = \lambda\mathcal{L}_{ce} + \beta \mathcal{L}_{seg},
\end{equation}
where $\alpha$ is set to 0.5, $\lambda$ to 2.0 and $\beta$ to 0.5.

\section{Experiment}
\label{sec:exp}
%







\begin{table*}[t!]
    \centering
    \caption{\textbf{Performance on referring segmentation benchmarks.} The evaluation metric is cIoU. ``ft" denotes fine-tuning on the specific dataset.}\vspace{-3mm}
    \resizebox{0.86\textwidth}{!}{
    \begin{tabular}{c|c|ccc|cc|ccc|ccc}
    \toprule[0.2em]
   \multirow{2}{*}{Method} & \multirow{2}{*}{LLM Size} & \multicolumn{3}{c|}{RefCOCO+} & \multicolumn{2}{c|}{RefCOCOg} & \multicolumn{3}{c|}{RefCOCO} & \multicolumn{3}{c}{gRefCOCO} \\
   ~ & ~ & val & testA & testB & val(U) & test(U) & val & testA & testB  & val & testA & testB \\
    \midrule
    \multicolumn{13}{c}{Referring Segmentation Specialist Without MLLM} \\
    \midrule
    VLT~\cite{ding2021vision} & - & 56.3 & 61.0 & 50.1 & 55.0 & 57.7 & 67.5 & 70.5 & 65.2 & 52.5 & 62.2 & 50.5 \\
    CRIS~\cite{wang2022cris} & - & 62.3 &  68.1 & 53.7 & 59.9 & 60.4 & 70.5 & 73.2 & 66.1 & 55.3 & 63.8 & 51.0 \\
    LAVT~\cite{yang2022lavt} & - & 62.1 & 68.4 & 55.1 & 61.2 & 62.1 & 72.7 & 75.8 & 68.8 &  57.6 & 65.3 & 55.0 \\
    PolyFormer-L~\cite{liu2023polyformer} & - & 69.3 & 74.6 & 61.9 & 69.2 & 70.2 & 76.0 & 78.3 & 73.3 & - & - & - \\
    ReLA~\cite{GRES} & - & 66.0 & 71.0 & 57.7 & 65.0 & 66.0 & 73.8 & 76.5 & 70.2 & 56.4 & 59.0 & 58.4 \\
    \midrule
    \multicolumn{13}{c}{MLLMs With Vision Expert} \\
    \midrule
    LISA (ft) ~\cite{lai2023lisa} & 7B  &  65.1 & 70.8 & 58.1 & 67.9 & 70.6 & 74.9 & 79.1 & 72.3 & - & - & -  \\
    PixelLM~\cite{ren2023pixellm} & 7B &  66.3 & 71.7 & 58.3 & 69.3 & 70.5 & 73.0 & 76.5 & 68.2 & - & - & -   \\
    GSVA (ft)~\cite{xia2023gsva} & 7B  & 64.5 & 67.7 & 58.6 & 71.1 & 72.0 & 76.4 & 77.4 & 72.8 &61.7 & 69.2 & 60.3 \\
    GroundHog~\cite{zhang2024groundhog} & 7B  & 70.5 & 75.0 & 64.9 & 74.1 & 74.6 &78.5 & 79.9 & 75.7 & 66.7 & - & - \\
    GlaMM (ft)~\cite{hanoona2023GLaMM} & 7B  & 72.6 & 78.7 & 64.6 & 74.2 & 74.9 & 79.5 & 83.2 & 76.9 & - & - & - \\
    SAM4MLLM~\cite{chen2024sam4mllm} & 7B  & 73.5 & 77.8 & 65.8 & 74.5 & 75.6 & 79.6 & 82.8 & 76.1 & 66.3 & 70.1 & 63.2 \\
    LaSagnA~\cite{wei2024lasagna} & 7B  & 66.4 & 70.6 & 60.1 & 70.6 & 71.9 &76.8 & 78.7 & 73.8 & 38.1 & 50.4 & 42.1 \\
    OMG-LLaVA (ft)~\cite{OMGLLaVA} & 7B & 69.1 & 73.1 & 63.0 & 72.9 & 72.9 & 78.0 & 80.3 & 74.1 & - & - & - \\
    F-LLM~\cite{wu2024flmm} & 7B & 65.8 & 75.2 & 58.5 & 70.1 & 71.7 & 75.8 & 79.5 & 72.4 & - & - & - \\
    Sa2VA~\cite{yuan2025sa2va} & 4B & 74.3 & - & - & 76.7 & - & 80.4 & - & - & - & - & -\\
    \midrule
    \multicolumn{13}{c}{MLLMs Without Vision Expert} \\
    \midrule
    Pixel-SAIL & 0.5B  & 70.8 & 75.8 & 65.4 & 75.4 & 76.7 & 77.9 & 80.5 & 75.9 & 63.9 & 71.5 & 63.6\\
    Pixel-SAIL (ft) & 0.5B  & 73.0 & 77.0 & 68.0 &75.6 & 76.1  &  79.1 & 81.7 & 77.0 & 68.0 & 74.0 & 66.8 \\
    Pixel-SAIL & 3B  &  75.7 & \textbf{79.7} & \textbf{72.0} & \textbf{78.7} & \textbf{80.4} & 80.8 & 82.6 & \textbf{79.0} & 67.7 & 74.6 & 67.1 \\
   Pixel-SAIL (ft) & 3B  & \textbf{76.2} & \textbf{79.7} & 71.2 & 78.5 & 79.4 & \textbf{81.8} & \textbf{83.4} & 78.8  & \textbf{72.1} & \textbf{77.1} & \textbf{70.4} \\
    \bottomrule[0.1em]
    \end{tabular}
    }
    \label{tab:res}
\end{table*}

\begin{table}[t!]
    \centering
    \vspace{-3mm}\caption{\textbf{Region caption performance on RefCOCOg dataset.}}\vspace{-2mm}
    \resizebox{0.46\textwidth}{!}{
    \begin{tabular}{c|cc|ccccc}
    \toprule[0.2em]
      Method & Pixel-SAIL & Pixel-SAIL & Sa2VA & OMG-LLaVA & Osprey & GLaMM \\
      Size & 0.5B & 3B & 4B & 7B & 7B & 7B \\
      \midrule
      METEOR & 16.0 & 17.6 & 17.3 & 15.3 & 16.6 & 16.2 \\
    \bottomrule[0.1em]
    \end{tabular}
    }
    \label{tab:region_cap}
\end{table}

\begin{table}[t!]
    \centering
    \vspace{-2mm}\caption{\textbf{The performance on our PerBench.} Due to the lack of visual prompt understanding capability, LISA scores 0 on all tasks.}\vspace{-3mm}
    \resizebox{0.46\textwidth}{!}{
    \begin{tabular}{c|c|c|c|cc|c}
    \toprule[0.2em]
      \multirow{2}{*}{Model} & \multirow{2}{*}{Size} & \multicolumn{1}{c|}{Detailed Caption} & \multicolumn{1}{c|}{MCQ} & \multicolumn{2}{c|}{V-T RES} & Overall \\
   ~ & ~ & METEOR & Acc & cIoU & gIoU & Score \\
    \midrule
    LISA~\cite{lai2023lisa} & 7B & 0 & 0 & 0 & 0 & 0 \\
    Osprey~\cite{yuan2023osprey} & 7B & 13.4 & 0.12 & 0 & 0 & 8.5\\
    GLaMM~\cite{hanoona2023GLaMM} & 7B & 12.6 & 0.14 & 24.3 & 14.6 & 15.3 \\
    Sa2VA~\cite{yuan2025sa2va} & 4B & 19.2 & 0.71 & 31.9 & 21.9 & 39.0  \\
    \midrule
    Pixel-SAIL & 0.5B & 21.4 &  0.69 & 29.7 & 19.8 & 38.4\\
    Pixel-SAIL & 3B & 24.2 & 0.74 & 33.4 & 23.5 & 42.2\\
    \bottomrule[0.1em]
    \end{tabular}
    }
    \label{tab:perbench}
\end{table}

\begin{table}[t!]
    \centering
    \vspace{-3mm}\caption{\textbf{Performance on the VQA benchmarks.} $\star$ refers to the use of an 800$^2$ resolution, which differs from the 1600$^2$ resolution in the pre-trained model.}\vspace{-2mm}
    \resizebox{0.46\textwidth}{!}{
    \begin{tabular}{c|c|cccc}
    \toprule[0.2em]
     Model & LLM Size & MME & MMBench & SEED & MMStar \\
     \midrule
     SOLO & 0.5B & 523.2/222.5 & 13.8 & 45.5 & 26.2 \\
     SOLO & 3B & 1155.7/257/5 & 53.4 & 65.4 & 40.3 \\
     EVEv2$\star$ & 7B & 1128.0/240.7 & 60.3 & 54.2 & 44.9 \\
     \midrule
     Pxiel-SAIL & 0.5B & 564.1/150.7 & 31.8 & 52.2 & 26.3  \\
     Pixel-SAIL & 3B & 1187.3/242.9 & 56.3 & 66.1 & 40.1 \\
     Pixel-SAIL$\star$ & 7B & 1081.0/260.4 & 58.9 & 64.7 & 44.3\\
    \bottomrule[0.1em]
    \end{tabular}
    }\vspace{-3mm}
    \label{tab:qa_bench}
\end{table}


\begin{table*}[t!]
    \begin{minipage}[c]{0.28\textwidth}
        \centering
        \caption{\textbf{Ablation study on the components of Pixel-SAIL.} ``RC" denotes region caption on RefCOCOg dataset.}\vspace{-2mm}
        \label{tab:component}
        \resizebox{0.98\textwidth}{!}{
        \begin{tabular}{l|c|c}
        \toprule[0.2em]
          Model & RefCOCO/+/g & RC \\
        \midrule
        Plain Baseline & 64.5/57.3/60.1 & 1.0 \\
        \hline
        + Upsampling & 69.7/62.5/65.3 & 0.9 \\
        + Training Data & 76.2/69.6/73.8 & 1.4 \\
        + VP Injection & 77.4/70.4/75.2 & 16.1 \\
        + Distillation & 77.9/70.8/75.4 & 16.0 \\
        \bottomrule[0.1em]
        \end{tabular}
        }  
    \end{minipage}\hfill
    \begin{minipage}[c]{0.24\textwidth}
        \centering
        \caption{\textbf{Ablation study on Base MLLM.} The training data only includes LLaVA-665k and RefCOCO/+/g.}\vspace{-2mm}
        \label{tab:base_mllm}
        \resizebox{0.98\textwidth}{!}{
        \begin{tabular}{c|c|c}
        \toprule[0.2em]
          MLLM & Size & RefCOCO/+/g \\
        \midrule
        SOLO & 0.5B &  69.7/62.5/65.3 \\
        SOLO & 3B & 73.2/66.4/69.1 \\
        EVEv2 & 7B &  74.9/68.7/71.3 \\
        \bottomrule[0.1em]
        \end{tabular}
        }
    \end{minipage}
    \begin{minipage}[c]{0.24\textwidth}
        \centering
        \caption{\textbf{Ablation on the training data.} ``RC" denotes region caption on RefCOCOg dataset.}
        \label{tab:data}
        \resizebox{0.98\textwidth}{!}{
        \begin{tabular}{c|c|c}
        \toprule[0.2em]
          Data & RefCOCO/+/g & RC \\
        \midrule
        Basic Data & 69.7/62.5/65.3 & - \\
        + Seg Data & 76.2/69.6/73.8 & - \\
        + VP Data & 77.4/70.4/75.2 & 16.1 \\
        \bottomrule[0.1em]
        \end{tabular}
        }
    \end{minipage}
    \begin{minipage}[c]{0.22\textwidth}
        \centering
        \caption{\textbf{Ablation study on the distillation strategy.}}\vspace{-1mm}
        \label{tab:distill}
        \resizebox{0.98\textwidth}{!}{
        \begin{tabular}{c|c}
        \toprule[0.2em]
          Data & RefCOCO/+/g \\
        \midrule
        w/o Distill & 77.5/70.5/75.5 \\
        M2F & 77.7/71.0/75.8  \\
        SAM2 & 77.8/70.9/75.9\\
        Both & 78.1/70.8/76.1\\
        \bottomrule[0.1em]
        \end{tabular}
        }
    \end{minipage}\vspace{-5mm}
\end{table*}

\noindent
\textbf{Implementation Details.} We extensively evaluate our meta-architecture using two open-source encoder-free multimodal large language models: SOLO~\cite{chen2024single} and EVEv2~\cite{diao2025evev2}. 
For SOLO, following~\cite{Sail2025}, we modify the attention mechanism between vision tokens from causal attention to full attention and conduct supervised fine-tuning on the LLaVA-1.5 665k dataset.
For SOLO, we modify the attention mechanism between vision tokens from causal attention to full attention and replace the LLM with Qwen2.5~\cite{yang2024qwen2} 0.5B and 3B, respectively.
For EVEv2, we retain its original architecture and weights without any modifications.
We build Pixel-SAIL 0.5B and 3B based on our modified SOLO baseline, and 7B on EVEv2. 
When training Pixel-SAIL based on SOLO, we maintain the original resolution of input images. 
For images with a long side exceeding 1024, we preserve the aspect ratio and resize the long side to 1024. 
When training Pixel-SAIL based on EVEv2, we resize the images to the closest to 800$^2$ pixels to reduce training costs, which differs from the original setting of 1600$^2$. 
The training process is conducted on 32 A100 (80GB) GPUs using the AdamW~\cite{ADAMW} optimizer with a cosine decay learning rate scheduler. 
We set the initial learning rate to 4e-5, the warm-up ratio to 0.03, and the batch size to 256. 
The training duration for the 0.5B and 3B models is 12 hours and 24 hours, respectively.

\noindent
\textbf{Evaluation Setup.} For visual prompt understanding and general image QA tasks, we adhere to the same setting as the base MLLM. In the case of segmentation-related tasks, if the model fails to predict a [SEG] token, we compel it to produce a [SEG] token to ensure the generation of the segmentation result.

\vspace{-3mm}\subsection{Main Results}
\label{exp:main_results}

\noindent
\textbf{Results on Referring Segmentation Benchmarks.}
We compare Pixel-SAIL with other pixel-grounded MLLMs and segmentation specialists on the RefCOCO+~\cite{yu2016modeling}, RefCOCOg~\cite{yu2016modeling}, RefCOCO~\cite{kazemzadeh2014referitgame}, and gRefCOCO~\cite{GRES} datasets. The comparison results are shown in Tab.~\ref{tab:res}. Pixel-SAIL 0.5B achieved 70.8, 75.4, and 77.9 cIoU on the validation splits of RefCOCO+, RefCOCOg, and RefCOCO, outperforming all segmentation specialists with comparable model sizes while also maintaining image conversation capabilities. Compared to the classical SAM-based MLLM competitor LISA-7B~\cite{lai2023lisa}, Pixel-SAIL 0.5B surpassed it by 4.2, 7.9, and 7.8 cIoU on RefCOCO, RefCOCO+, and RefCOCOg respectively, despite having a much smaller model size (0.5B \textit{vs.} 7B). On the more complex gRefCOCO dataset that includes multi-object segmentation, Pixel-SAIL 0.5B outperformed the carefully designed GSVA-7B~\cite{xia2023gsva} by 6.3, 4.8, and 6.5 cIoU on validation, testA, and testB splits respectively.

When scaling the model to 3B, Pixel-SAIL achieved 75.7, 78.7, 80.8, and 67.7 cIoU on RefCOCO+, RefCOCOg, RefCOCO, and gRefCOCO datasets respectively, surpassing all larger-sized (7B) MLLMs assisted with vision experts. Pixel-SAIL-3B even outperformed the SOTA Sa2VA-4B~\cite{yuan2025sa2va} (which uses the powerful InternVL2-4B~\cite{chen2024expanding} and SAM2-L~\cite{ravi2024sam2}), achieving performance advantages of 1.4 and 2.0 cIoU on the more challenging RefCOCO+ and RefCOCOg datasets respectively.


\noindent
\textbf{Results on Visual Prompt Understanding Benchmarks.} We evaluate the region caption performance on the RefCOCOg dataset, with results shown in Tab.~\ref{tab:region_cap}. The training dataset of Pixel-SAIL does not include the RefCOCOg region caption dataset, so we directly evaluate its zero-shot performance. Pixel-SAIL-0.5B achieves a METEOR score of 16.0, surpassing OMG-LLaVA 7B by 0.7 points. When scaling the model to 3B, Pixel-SAIL achieves a METEOR score of 17.6, outperforming carefully designed larger models such as Osprey 7B and GLaMM 7B by 1.0 and 1.4 points respectively.

\noindent\textbf{Results on PerBench.} We have benchmarked several popular pixel-grounded MLLMs on our proposed PerBench, with results shown in Tab.~\ref{tab:perbench}. LISA~\cite{lai2023lisa} scores 0 points across all tasks due to its inability to understand visual prompt inputs. Osprey~\cite{yuan2023osprey} demonstrates strong object caption capabilities; however, it achieved only 13.4 METEOR in detailed caption tasks and 12.0\% accuracy in MCQ tasks due to limitations from short object caption lengths in its training data and impaired instruction-following ability. GLaMM~\cite{hanoona2023GLaMM} and Sa2VA~\cite{yuan2025sa2va} both exhibit comprehensive prompt understanding and segmentation capabilities, though GLaMM's weaker instruction-following ability resulted in only 14.0\% accuracy in MCQ tasks. Pixel-SAIL-0.5B achieves an overall score of 38.4, comparable to Sa2VA-4B despite Pixel-SAIL having a more powerful base MLLM and segmentation expert. Notably, Pixel-SAIL-3B achieves an overall score of 42.2, outperforming Sa2VA-4B across all three tasks.

\noindent
\textbf{Results on VQA Benchmarks.}
We compare the visual question answering performance of Pixel-SAIL with the corresponding base MLLMs on the MME~\cite{fu2023mme}, MMBench~\cite{liu2023mmbench}, SEED~\cite{li2023seed}, and MMStar~\cite{chen2024mmstar} benchmarks, and the results are presented in Tab.~\ref{tab:qa_bench}. When the model size is 0.5B, Pixel-SAIL demonstrates performance improvements over the base MLLM across all four benchmarks, particularly on MMBench, where the score increased from 13.8 to 31.8. However, when the model size is 3B and 7B, Pixel-SAIL's performance is on par with that of the base MLLMs, which may be constrained by the current quantity (less than 2M) and quality of visual prompts and segmentation data.

\begin{figure}
  \centering
  \includegraphics[width=0.98\linewidth]{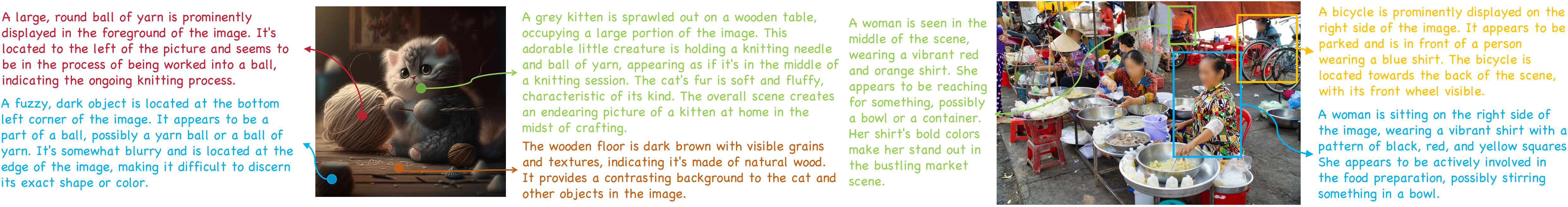}
  \includegraphics[width=0.98\linewidth]{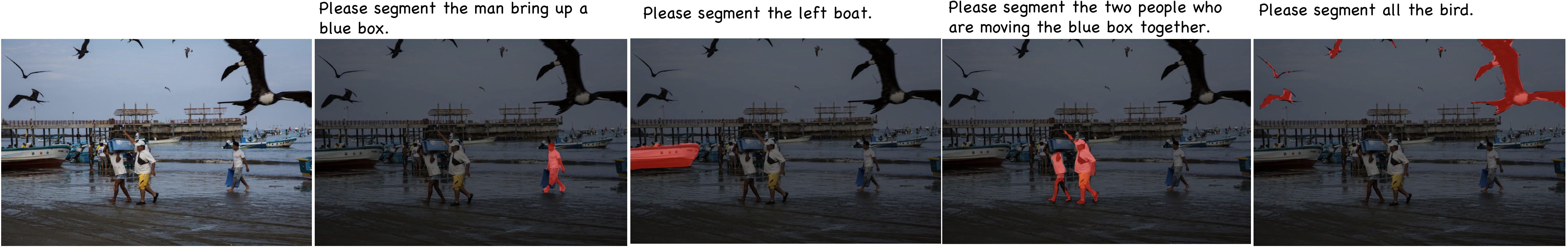}
  \includegraphics[width=0.98\linewidth]{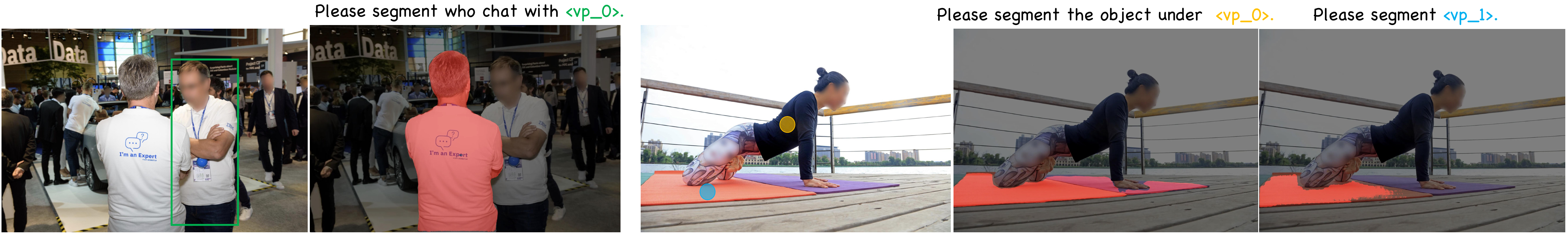}
  \includegraphics[width=0.98\linewidth]{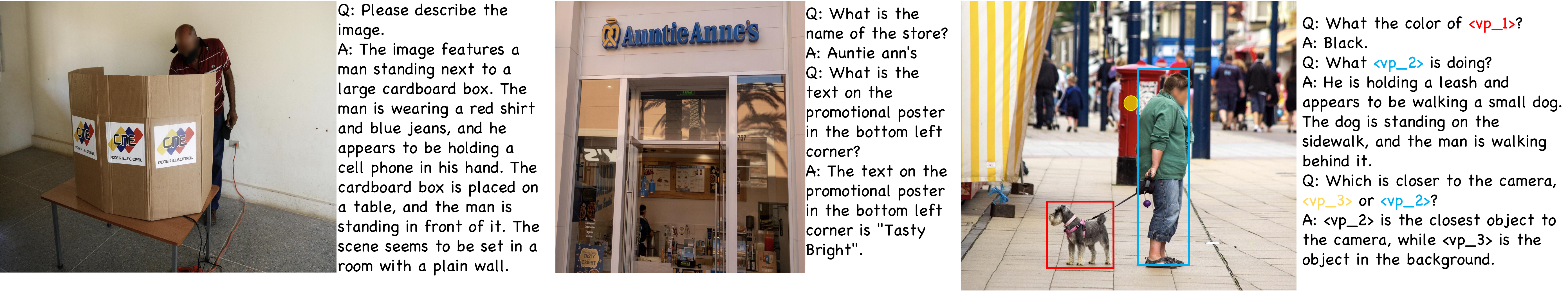}
  \vspace{-3mm}\caption{\textbf{Visualization results of Pixel-SAIL on diversity tasks. Best view it in color and zoom in.} From top to bottom are visual prompt-based object caption, single/multi-object referring segmentation, vision-text referring segmentation, image caption and QA, and visual-prompt based conversation. Visual prompts in the form of points and boxes are converted into mask prompts using SAM~\cite{kirillov2023segment}. For more visualization results and comparisons with other MLLMs, please refer to the appendix.}\vspace{-3mm}
  \label{fig:demos}
\end{figure}

\begin{figure}
  \centering
  \includegraphics[width=0.98\linewidth]{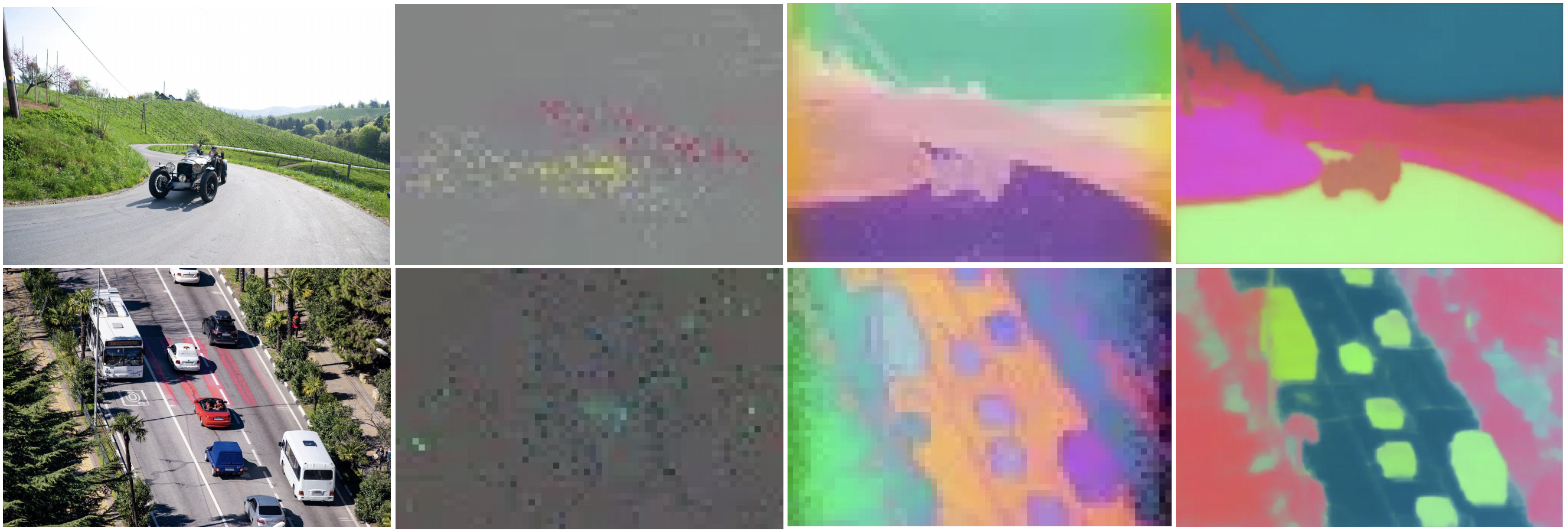}
  \includegraphics[width=0.98\linewidth]{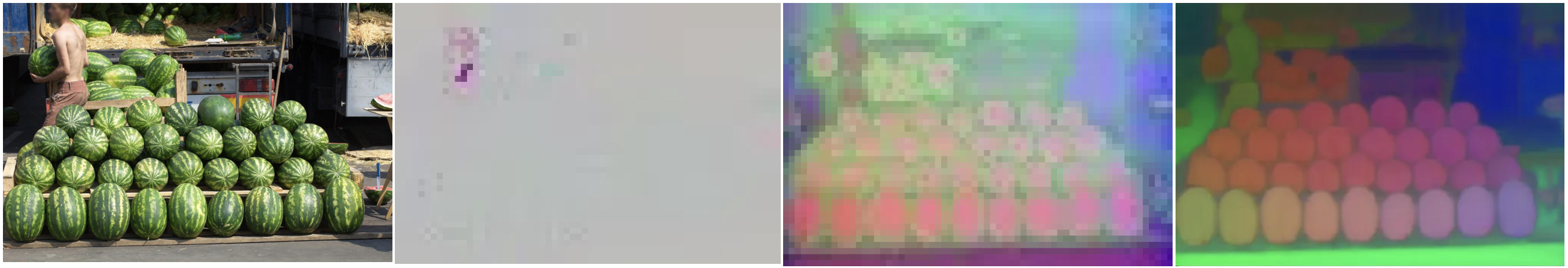}
  \vspace{-2mm}\caption{\textbf{Image feature visualization results.} From left to right are the image feature of the base MLLM, the image feature of Pixel-SAIL, and the mask feature of Pixel-SAIL.}\vspace{-5mm}
  \label{fig:feat_vis}
\end{figure}

\subsection{Ablation Studies}
\label{exp:ablation}

\noindent
\textbf{Effectiveness of Each Component.}
We conduct comprehensive ablation studies on the proposed components, with results presented in Tab.~\ref{tab:component}. Our plain baseline, trained with LLaVA-665k and RefCOCO/+/g data, achieves only 64.5, 57.3, and 60.1 cIoU on the RefCOCO, RefCOCO+, and RefCOCOg datasets, respectively. Moreover, this baseline completely fails on the visual prompt understanding task, attaining merely 1.0 METEOR on the region caption task.
Upon incorporating the learnable upsampling module, segmentation quality improves dramatically, with the model reaching 76.2, 69.6, and 73.8 cIoU on RefCOCO, RefCOCO+, and RefCOCOg. However, the model still cannot effectively interpret user-input visual prompts due to insufficient semantic information in the object representation.
When we scale up the training data by introducing substantial amounts of segmentation data and visual-prompt understanding data, the model's segmentation capabilities are further enhanced. Despite scaling the training data, the model continues to struggle with visual prompt inputs because of the limited semantic information in the object representation.
After implementing our proposed visual prompt injection mechanism, the model demonstrates significant improvements in visual prompt understanding, achieving 16.1 METEOR on the region caption task. Interestingly, we observe that enhanced visual prompt understanding capabilities positively influence referring segmentation performance. Finally, incorporating the distillation strategy further refines the model's detailed segmentation quality.

\noindent
\textbf{Ablation on Various MLLMs.}
To demonstrate the effectiveness of Pixel-SAIL, we validate across different architectures and sizes, with results shown in Tab.~\ref{tab:base_mllm}. To reduce training costs, we use only LLaVA-665k and RefCOCO/+/g data for training and evaluate on the referring segmentation task. When using our modified 0.5B SOLO as the base MLLM, Pixel-SAIL achieves cIoU scores of 69.7, 62.5, and 65.3 on RefCOCO/+/g. When scaling the model size to 3B, Pixel-SAIL's performance improves by 3.5, 3.9, and 3.8 cIoU on RefCOCO/+/g. When using EVEv2-7B as the base MLLM, despite the attention between vision tokens changing from full attention to causal attention and the architecture transitioning to an MOE architecture, Pixel-SAIL achieves cIoU scores of 77.4, 70.4, and 75.2 on RefCOCO/+/g, demonstrating that performance consistently increases with model scaling.

\noindent
\textbf{Ablation on Data Scaling.}
Data plays a crucial role in the performance of Pixel-SAIL. As showin in Tab.~\ref{tab:data}, we conduct comprehensive ablation studies on the training data to evaluate its impact. When trained solely with basic data (including LLaVA-665k and RefCOCO/+/g datasets), Pixel-SAIL achieves 69.7, 62.5, and 65.3 cIoU on RefCOCO, RefCOCO+, and RefCOCOg, respectively. Upon scaling the segmentation-related data, Pixel-SAIL demonstrates significant performance improvements of 6.5, 7.1, and 8.5 cIoU on these datasets. Furthermore, incorporating visual prompt data for mixed training not only enhances the model's visual prompt understanding capabilities but also yields additional performance gains of 1.2, 0.8, and 1.4 cIoU on RefCOCO, RefCOCO+, and RefCOCOg, respectively.

\noindent
\textbf{Ablation on Distillation Strategy.}
Distillation is a highly effective method for infusing knowledge into Pixel-SAIL. We conduct ablation studies on the distillation strategy, and the results are presented in Tab.~\ref{tab:distill}. We use the average cIoU across all splits as the evaluation metric. When only Mask2Former~\cite{cheng2021mask2former} is employed to distill high-resolution mask features, Pixel-SAIL achieves performance gains of 0.2, 0.5, and 0.3 on RefCOCO/+/g. When SAM2~\cite{ravi2024sam2} is used to distill low-resolution image features, Pixel-SAIL obtains performance improvements of 0.3, 0.4, and 0.4 on RefCOCO/+/g. When both teacher models are utilized collaboratively, performance gains of 0.6, 0.3, and 0.5 are achieved. Additionally, the extra computational cost introduced by the distillation strategy is minimal, increasing the training time by only about 5\% for Pixel-SAIL-0.5B.

\subsection{Visualization Analysis}
\label{exp:visual_results}

\noindent
\textbf{Visual Comparison.} In Fig.~\ref{fig:demos}, we showcase Pixel-SAIL's visualization results on diverse tasks. Pixel-SAIL flexibly interprets both visual prompts and text instruction inputs, responding with text and segmentation masks.

\noindent
\textbf{Visual Affinity Map Analysis.} We use PCA dimensionality reduction algorithm to visualize vision features, with results shown in Fig.~\ref{fig:feat_vis}. Our Pixel-SAIL's image features (3rd column) are denser and more diverse compared to the base MLLM's image features (2nd column). Pixel-SAIL's mask features, after the upsampling module, are denser and have better segmentation edges. Interestingly, Pixel-SAIL's image features (more focused on understanding, combining factors such as categories, colors, positions, etc.) exhibit different characteristics from mask features (more focused on perception, categories, and instances). As seen in the second row's third and fourth columns, the cars on the left and right have relatively distant feature representations in the image features, while they are very close in the mask features.

\section{Conclusion}
\label{sec:conclusion}

We explore the simplest architecture for pixel-grounded understanding tasks. In particular, we present Pixel-SAIL, which extends current SAIL-like MLLM for fine-grained understanding with three technical improvements (learnable upsampling module, new visual prompt encoding , and segmentor feature distillation). For the first time, our work proves that even without extra visual experts (visual encoder, segmentation models), one single transformer can still achieve stronger performance on four public referring segmentation benchmarks.
We further introduce a more challenging benchmark, Perbench, to promote the development of pixel-MLLM community.

\noindent
\textbf{Limitation and Future Work.} Our work provides the simplest solution for pixel-grounded tasks. However, one limitation is that we only adopt 1.7M data for co-training. We will further explore Pixel-SAIL on more data (for example, billion-level masks along with visual prompts~\cite{kirillov2023segment}) for co-training.
\clearpage
\newpage
{
    \small
    \bibliographystyle{ieeenat_fullname}
    \bibliography{main}
}

\clearpage
\setcounter{page}{1}
\maketitlesupplementary

We first present more details on training and testing of our Pixel-SAIL in Sec.~\ref{sub:trainig_testing}. Then, we present the detailed benchmark building process, in Sec.~\ref{sub:more_detailed_benchmark} and more challenging examples in PerBench in Sec.~\ref{sub:challenging_cases_PerBench}. Next, we present more comparison with current state-of-the-art pixel-grounded MLLMs, in Sec.~\ref{sub:more_comparison_pixel_grouned_MLLM}.

\section{More Detailed Training and Testing}
\label{sub:trainig_testing}

\noindent
\textbf{Training.} We will present more details about the training, including dataset sampling specifications and distillation methodology. For the RefCOCO series~\cite{kazemzadeh2014referitgame, yu2016modeling} datasets, we randomly sample 5 referring expressions per image and organize them into a multi-round dialogue format as a single training data point, processing all images for four epochs. For COCO~\cite{coco_dataset} data, we sample 5 categories per image and randomly select either instance mode or semantic mode to structure the responses. In instance mode, objects are arranged by their center points from left to right. We process the COCO dataset for one epoch. For Pixel2Cap~\cite{you2025pix2cap}, our generated detailed object caption data, and Osprey~\cite{yuan2023osprey} object description data, we randomly sample 1-5 visual prompts per image and randomly incorporate questions about non-existent visual prompts, with responses indicating that these visual prompts do not exist. These object caption datasets are processed for five epochs. For other segmentation-related or visual prompt-related data, we conduct one epoch. For LLaVA-665k, we randomly sample at a 1:1 ratio alongside other data for joint training to ensure that the base MLLM's instruction-following capability remains intact.

When the length of input tokens (including the length of vision tokens) exceeds 8192, we truncate the excess portion. For the 0.5B model, we use DeepSpeed Zero-1~\cite{rasley2020deepspeed} for training, and for the 3B and 7B models, we use DeepSpeed Zero-2~\cite{rasley2020deepspeed} for training.

We distill the mask features generated by the Mask2Former~\cite{cheng2021mask2former} pixel decoder and the lowest resolution features generated by the SAM2~\cite{ravi2024sam2} image encoder onto the upsampled mask features from Pixel-SAIL and the image features directly reshaped from vision tokens, respectively. We use bilinear interpolation to align spatial dimensions and implement a learnable linear layer to align the channel size. The distillation process employs MSE loss with a weight of 0.5.

\noindent
\textbf{Testing.} We have elaborated on the testing details of Pixel-SAIL on pixel-grounded benchmarks in the main text. For general image question answering benchmarks, we follow the prompt settings of the base MLLMs and use VLMEvalKit~\cite{duan2024vlmevalkit} for evaluation, without using additional LLM assistance to identify answers.

\begin{figure*}
  \centering
  \includegraphics[width=0.98\linewidth]{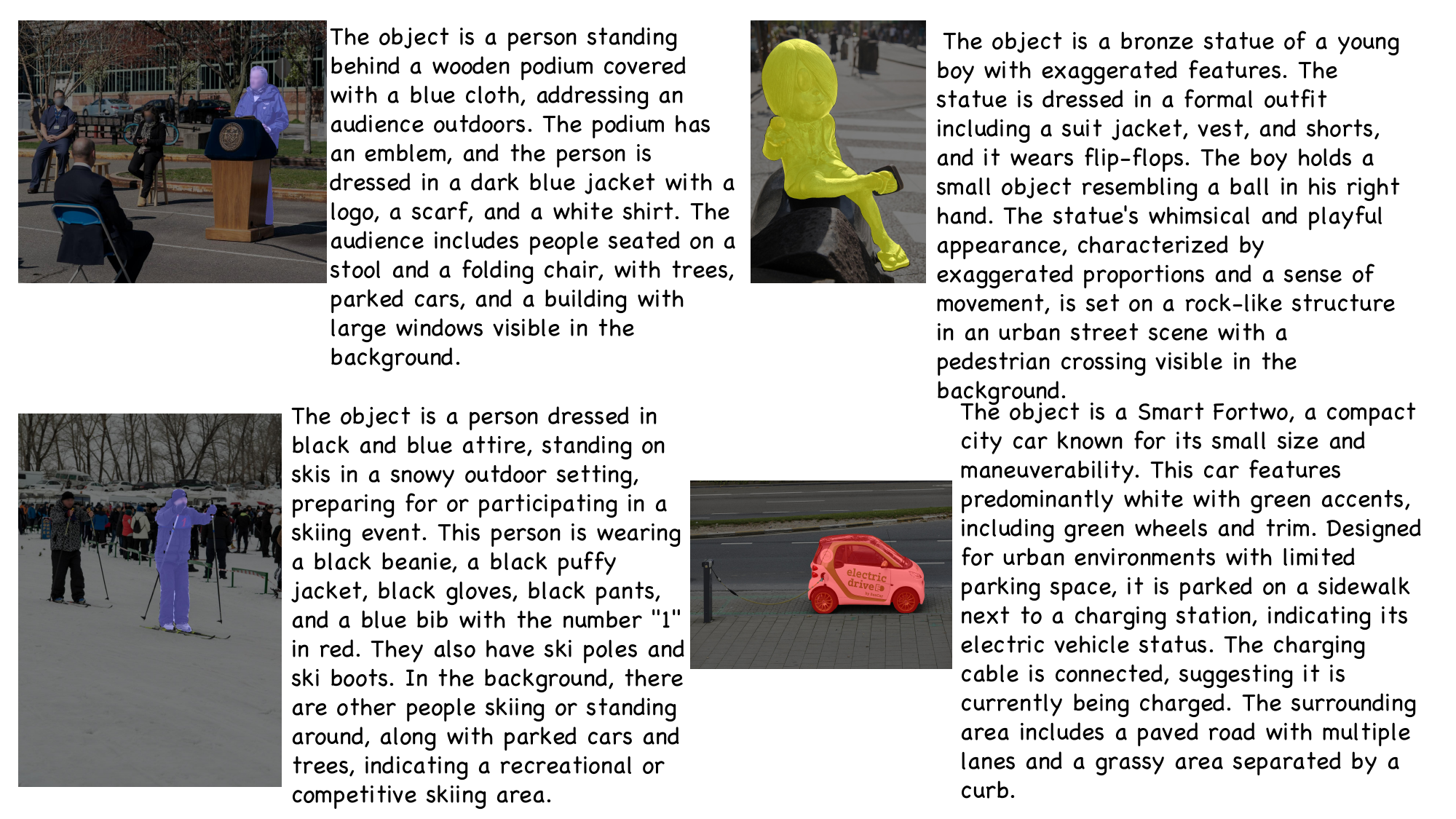}
  \includegraphics[width=0.98\linewidth]{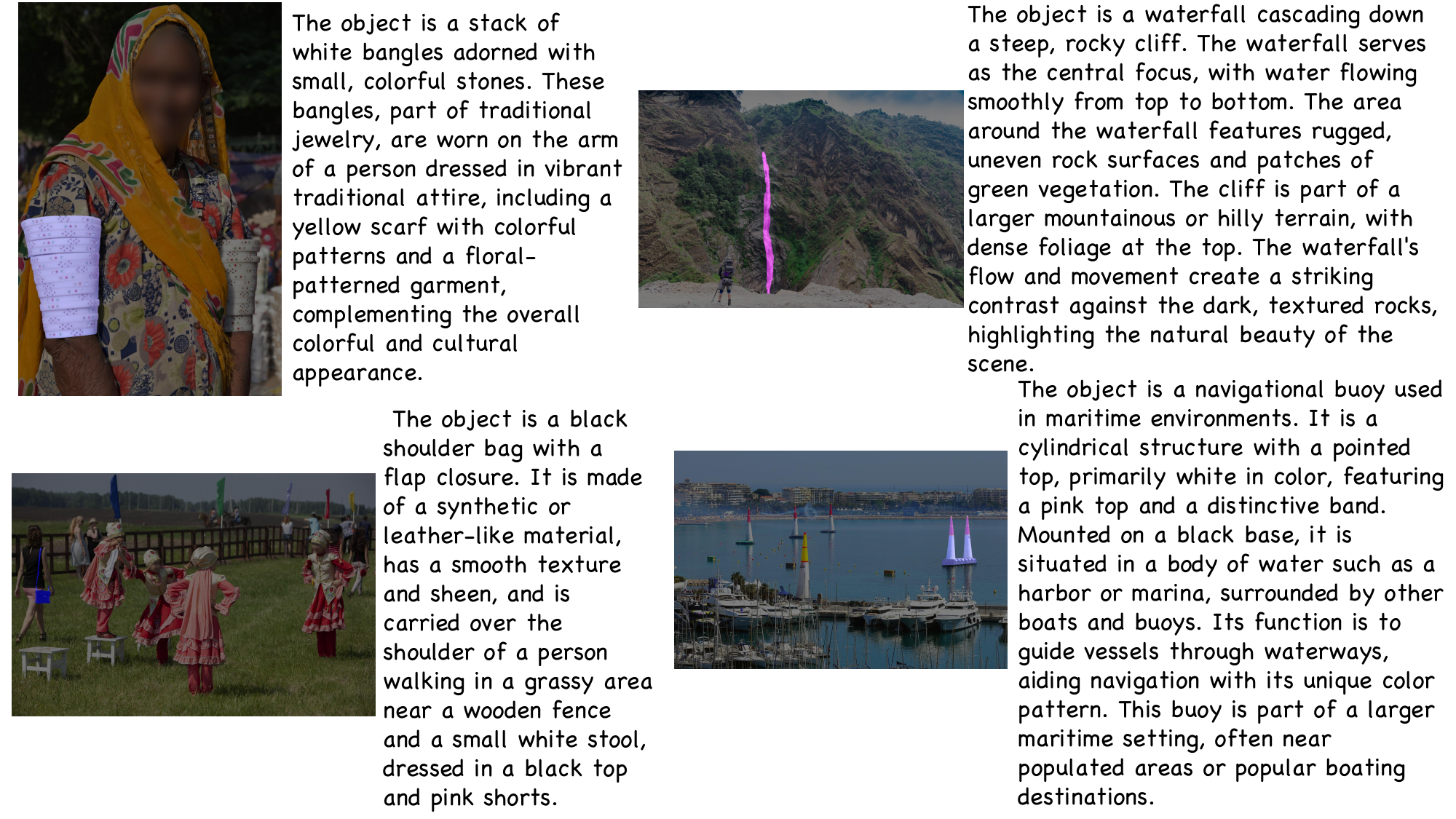}
  \caption{\textbf{More visualization examples of detailed object captions from our PerBench.}}
  \label{fig:more_perbench_detail_cap}
\end{figure*}

\begin{figure*}
  \centering
  \includegraphics[width=0.98\linewidth]{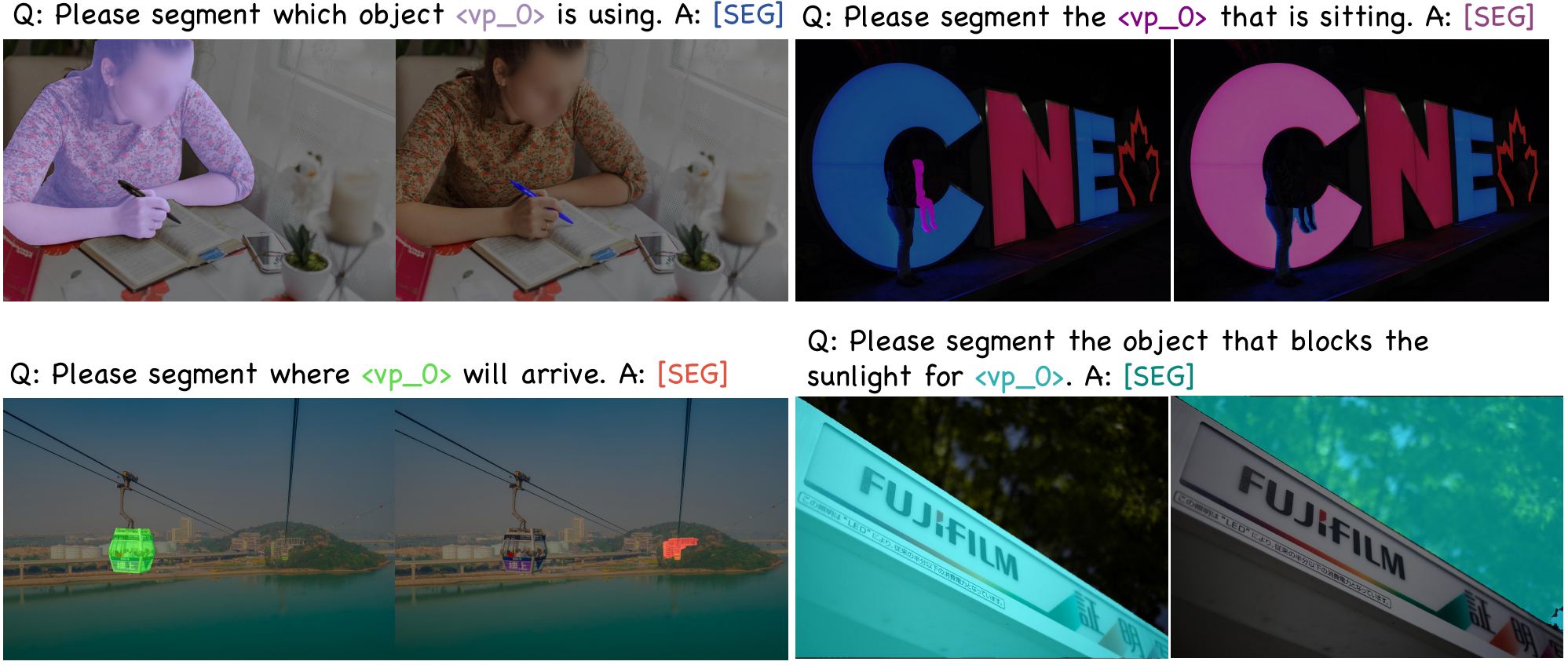}
  \includegraphics[width=0.98\linewidth]{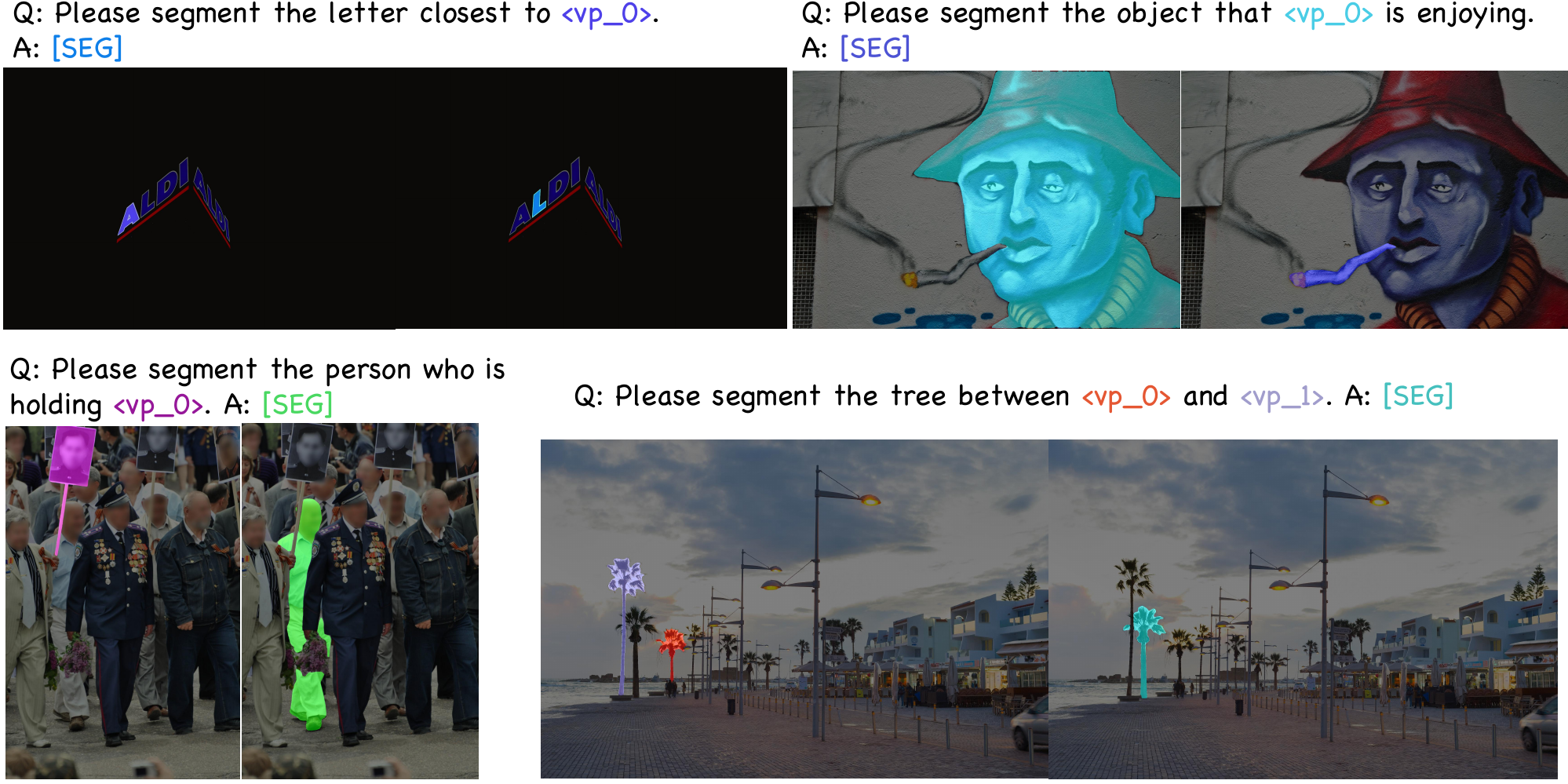}
  \caption{\textbf{More visualization examples of vision-text referring segmentation from our PerBench.}}
  \label{fig:more_perbench_vtres}
\end{figure*}

\section{More Detailed Process on Benchmarking Building}
\label{sub:more_detailed_benchmark}

The construction of PerBench combines an automated model-generated pipeline with manual screening, correction, and annotation. 
The process is divided into three stages.

The \textbf{first stage} involves annotating detailed object captions. We crop objects and draw visual prompts on the original images to prompt InternVL-2.5 78B~\cite{chen2024expanding} and Qwen-VL 2.5 72B~\cite{bai2025qwen2} to generate detailed captions for the objects. 
These captions are then cross-validated using Qwen2.5 72B~\cite{yang2024qwen2}. 
If all captions are consistent, they are integrated using an LLM; otherwise, the data are discarded. 
After the model automatically generates the detailed object captions, we manually select and correct 500 of them to form the final 500 detailed object caption data points in the benchmark.

The \textbf{second stage} focuses on annotating visual-prompt question-answering data in an MCQ (Multiple Choice Question) format. 
In this phase, we manually generate a multiple-choice question for each object caption obtained from the first stage. 
After completing the annotations, two quality control specialists perform cross-verification to identify and rectify any potential errors.

The \textbf{final stage} contains the annotation of visual-text referring segmentation data. At this stage, we manually select and annotate object segmentation masks, referring visual prompts, and text from SAM images. 
During the annotation process, we consider various factors such as positional relationships, event relationships, appearance, size, and more, including cases with both single and multiple visual prompts. 
Once the annotation is complete, two individuals review it, and correct the errors.

\section{More Challenging Cases in PerBench}
\label{sub:challenging_cases_PerBench}

We present more detailed object caption samples from our PerBench in Fig.~\ref{fig:more_perbench_detail_cap}. The objects are derived from diverse senses and categories, encompassing humans, man-made objects, and natural landscapes. 
The object captions include basic categories, attributes, purposes, and relationships with surrounding objects. 
This high-quality benchmark will effectively advance the development of the visual prompt understanding community.

More referring segmentation samples are illustrated in Fig.~\ref{fig:more_perbench_vtres}. Our manually annotated samples cover a variety of scenes, such as indoor and outdoor settings, and include objects of multiple granularities. 
The referring text encompasses positional relationships, event relationships, and more. This new task is more challenging than current pure text referring segmentation tasks.

\begin{figure*}
  \centering
  \includegraphics[width=0.98\linewidth]{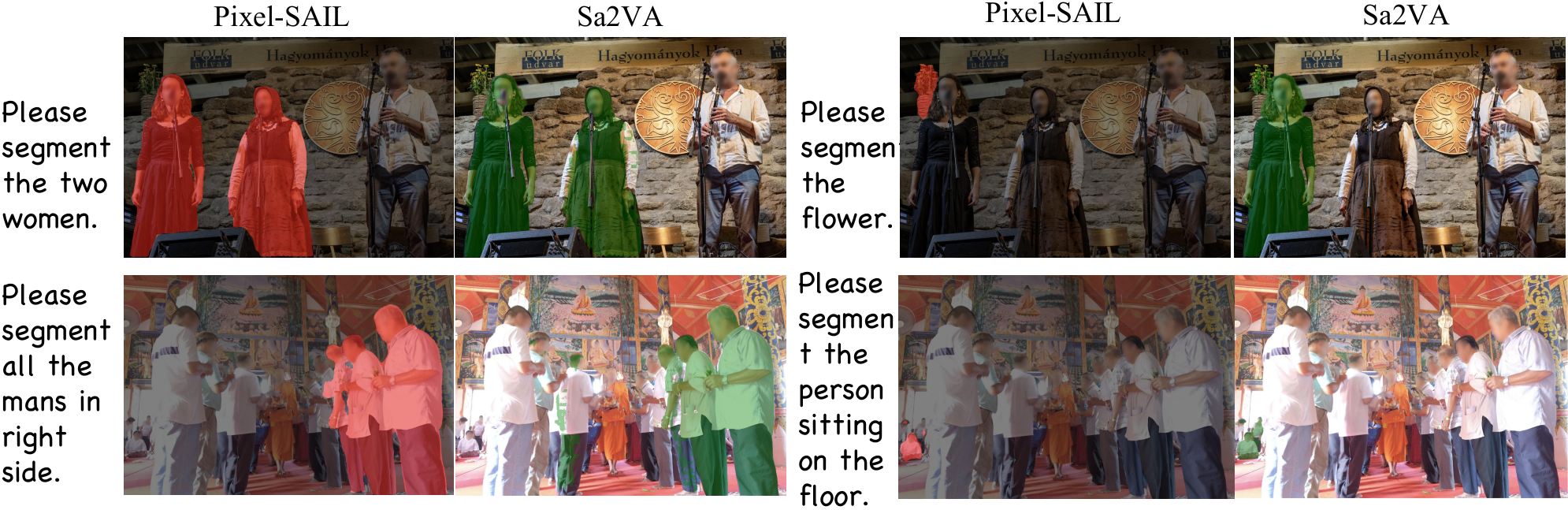}
  \caption{\textbf{Visualization Comparison of Sa2Va and Pixel-SAIL.}}
  \label{fig:compare_sa2va}
\end{figure*}

\section{More Comparison With SOTA Pixel-Grounded MLLM}
\label{sub:more_comparison_pixel_grouned_MLLM}

We conduct a qualitative comparative analysis with the SOTA pixel-grounded MLLM, Sa2VA~\cite{yuan2025sa2va}, and present the visualization results in Fig.~\ref{fig:compare_sa2va}. We observe that both Pixel-SAIL and Sa2VA achieve excellent results in most cases. However, Sa2VA performs significantly weaker than Pixel-SAIL in certain scenarios, despite utilizing the much more powerful InternVL2.5~\cite{chen2024expanding} compared to our base encoder-free MLLM~\cite{chen2024single}. In the left examples, Sa2VA performs notably worse than Pixel-SAIL in multi-object segmentation tasks. Additionally, in the right example, Sa2VA demonstrates significantly weaker attention to non-core areas of the image, such as edges, compared to Pixel-SAIL, leading to frequent failures in segmenting objects near image boundaries.


\end{document}